\documentclass{article}

\usepackage[nonatbib,preprint]{neurips_data_2024}
\usepackage{comment}

\usepackage[utf8]{inputenc} 
\usepackage[T1]{fontenc}    
\usepackage{hyperref}       
\usepackage{url}            
\usepackage{booktabs}       
\usepackage{amsfonts}       
\usepackage{nicefrac}       
\usepackage{microtype}      
\usepackage{xcolor}         
\usepackage{xspace}
\usepackage[normalem]{ulem}

\newcommand{\benchmark}{\textsc{WildHallucinations}\xspace}
\newcommand{\factscore}{\textsc{FActScore}\xspace}
\newcommand{\wildfactscore}{\textsc{WildFActScore}\xspace}
\newcommand{\wildfactscoreStrict}{\textsc{WildFActScore-Strict}\xspace}
\usepackage{todonotes}
\usepackage{graphicx}
\usepackage{pifont}
\usepackage{amsmath}
\usepackage{caption}
\usepackage{listings}
\usepackage{bbm}
\usepackage{multirow, makecell}
\newcommand{\cmark}{{\color{green}\ding{51}}}%
\newcommand{\xmark}{{\color{red}\ding{55}}}%
\usepackage[capitalize,noabbrev]{cleveref}

\title{\benchmark: Evaluating Long-form Factuality in LLMs with Real-World Entity Queries}

\author{%
 Wenting Zhao$^{1,3}$,
 Tanya Goyal$^{1}$,
 Yu Ying Chiu$^{2}$,
 Liwei Jiang$^{2,3}$, \\ \textbf{Benjamin Newman$^{2}$, Abhilasha Ravichander$^{3}$, Khyathi Chandu$^{3}$\hspace{0.1em},}\\ \textbf{Ronan Le Bras$^{3}$, Claire Cardie$^{1}$, Yuntian Deng$^{3,4}$, Yejin Choi$^{2,3}$} \\[1ex]
$^{1}$Cornell University \quad $^{2}$University of Washington \\ $^{3}$Allen Institute for Artificial Intelligence \quad 
$^{4}$University of Waterloo\\[1ex]
\texttt{\href{mailto:wz346@cornell.edu}{wz346@cornell.edu}}
}

\begin{document}

\maketitle

\begin{abstract}

While hallucinations of large language models (LLMs) prevail as a major challenge, existing evaluation benchmarks on factuality do not cover the diverse domains of knowledge that the real-world users of LLMs seek information about. To bridge this gap, we introduce \benchmark, a benchmark that evaluates factuality. It does so by prompting LLMs to generate information about entities mined from user-chatbot conversations in the wild. These generations are then automatically fact-checked against a systematically curated knowledge source collected from web search. Notably, half of these real-world entities do not have associated Wikipedia pages. We evaluate 118,785 generations from 15 LLMs on 7,919 entities. We find that LLMs consistently hallucinate more on entities without Wikipedia pages and exhibit varying hallucination rates across different domains. Finally, given the same base models, adding a retrieval component only slightly reduces hallucinations but does not eliminate hallucinations.
\end{abstract}

\section{Introduction}
Large language models (LLMs) have made significant progress in generating coherent texts. Despite these advancements, ensuring the factual accuracy of the generated content remains a formidable challenge~\cite{min2023factscore,mishra2024finegrained,hong2024hallucinations}. Hallucinations---instances where LLMs generate information that is unsupported or incorrect---pose a major obstacle to these models' reliable and safe deployment, particularly in high-stake applications~\cite {yan2024worse}. This issue becomes more serious as users trust the plausible-looking outputs from advanced LLMs~\cite{Choudhury_2024}.

\begin{figure}[!t]
    \centering
    \includegraphics[trim={0 15cm 0 0},clip,width=\textwidth]{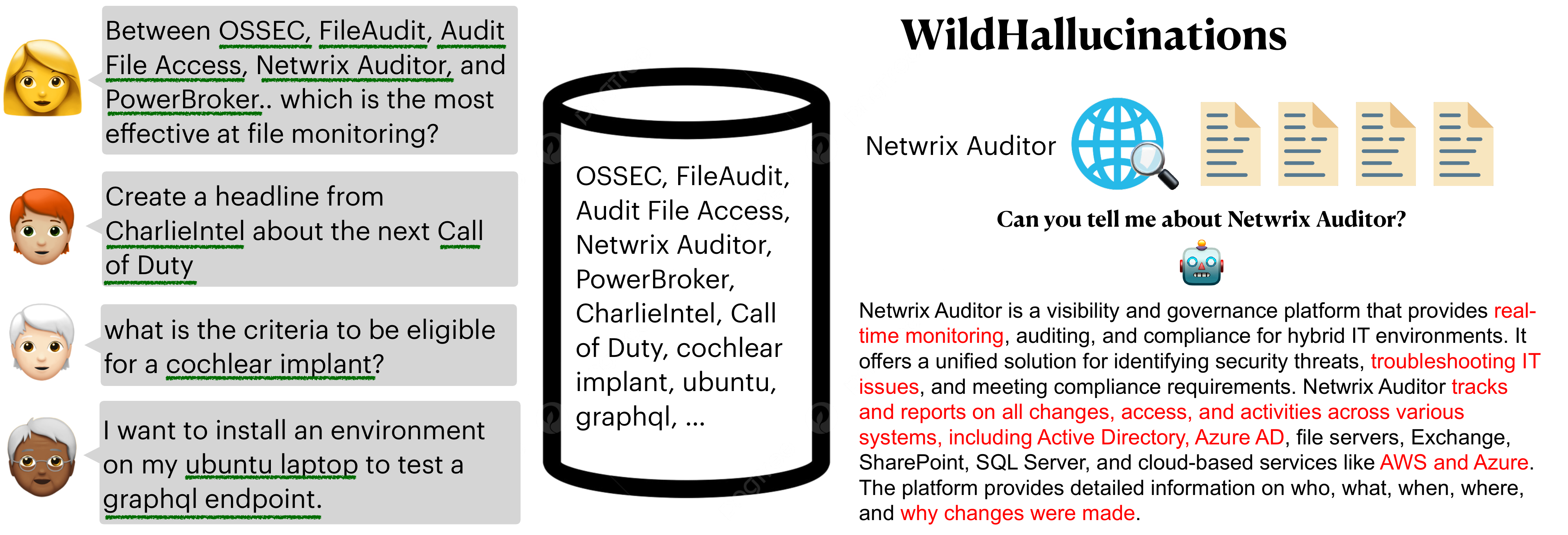}
    \caption{Overview of \benchmark. We mine entities from real-world user queries and construct a knowledge source for these entities using web search. We prompt LLMs to generate information about the entities and automatically fact-check the outputs using the knowledge source. The texts colored in red are hallucinated by an LLM.}
    \label{fig:overview}
\end{figure}

To understand the hallucination behaviors of LLMs in real-world use cases and to assist model developers in creating more reliable systems, we introduce \benchmark, a benchmark designed to evaluate the factuality of LLMs using entities from diverse domains such as computing, culture, finance, and more, collected from real-world user-chatbot interactions. \Cref{fig:overview} shows an overview of \benchmark. To construct the benchmark, we extract entities from the WildChat dataset~\cite{zhao2024wildchat}, which comprises one million user-chatbot interactions in the wild. Notably, 52\% of the extracted entities do not have corresponding Wikipedia pages, highlighting that users often seek information beyond the scope of Wikipedia. To evaluate the factuality of LLMs, we prompt them to generate descriptive texts about each entity. We then identify hallucinations in these generated descriptions using \factscore~\cite{min2023factscore}, an automatic fact-checking method for free-text generations. By evaluating LLM outputs for these entities, we extend factuality evaluation to cover diverse domains and non-Wikipedia knowledge. \Cref{tab:datasets} summarizes how \benchmark is related to and different from existing evaluation benchmarks.

\begin{table}[!t]
    \centering
    \caption{\label{tab:datasets}Comparison against existing hallucination benchmarks.{\protect\footnotemark} Among these, \factscore only caters to a limited domain, HaluEval-Wild does not contain reliable evidence documents for evaluation. ExpertQA consists of high-quality and challenging queries, but automatic evaluators report poor performance on this dataset. Our benchmark addresses these shortcomings by carefully curating reliable evidence documents and a reliable evaluation pipeline.}
    \small
    \begin{tabular}{@{}c|clll@{}}
    \toprule
       \textbf{Dataset}  &  \textbf{\#Examples} & \textbf{Query} & \textbf{Evidence Document} & \textbf{Domain} \\ \midrule

       ExpertQA \cite{malaviya2024expertqa} &	2,177 &	Experts & retrieval via Google	& Medicine, Law, etc. \\
HaluEval-Wild \cite{zhu2024halueval} &	6,505 &	ShareGPT &	GPT-4 reference &	Web queries	 \\
\factscore \cite{min2023factscore} &	500	& Person Bios	& Wikipedia article	& Wikipedia \\ \midrule

     Ours  &  \textbf{7,917} & WildChat entities & pre-retrieved via Google & \multirowcell{2}{Geography, Science \\ Person, Finance, etc.}\\
     & \\
    \bottomrule
    \end{tabular}
\end{table}

\footnotetext{We only include benchmarks that can evaluate free-text generations in this table. Another type of hallucination dataset contains fixed generation and pre-annotated factuality labels for given queries. These latter datasets are designed to evaluate a different LLM capability, i.e. error detection, and cannot be directly used for our task.}

We evaluate a diverse set of state-of-the-art LLMs on \benchmark, including standard LLMs and retrieval-augmented generation (RAG) models. Our findings reveal that: (1) LLMs exhibit varying hallucination rates across different domains, with higher rates in the people and finance domains, and lower rates in geographic and computing-related domains; (2) LLMs consistently hallucinate more on entities without Wikipedia pages compared to those with them; and (3) retrieval helps LLMs reduce hallucinations to some extent, but it is not sufficient to eliminate them entirely. We make \benchmark publicly available on Hugging Face\footnote{\url{https://huggingface.co/datasets/wentingzhao/WildHallucinations}} under the MIT license. We will regularly update the benchmark as more WildChat conversations become available, providing researchers and practitioners with a tool to evaluate the factual precision of LLMs.

\section{The \benchmark Benchmark}

We design our benchmark for evaluating hallucinations with two primary goals: (1) ensuring the evaluation process is both automatic and reliable, and (2) covering diverse types of information that real-world users seek. To meet both goals, we focus on testing LLMs' knowledge of entities. Compared to fact-checking answers to open-ended questions, verifying responses about entities is more objective and allows for the creation of a comprehensive knowledge source about these entities. We posit that mastering the entities included in user queries is a prerequisite for delivering factual responses. 
Furthermore, automatic evaluation for fact-checking LLMs' responses about entities is also available, as described in \factscore~\cite{min2023factscore}. 


\subsection{\label{sec:data}Dataset Creation}
The overall pipeline is outlined in \Cref{fig:overview}. The entities in our benchmark are extracted from WildChat \cite{zhao2024wildchat}, a dataset of real-world user-chatbot conversations\footnote{WildChat is released under the ODC-By license.}. For each extracted entity, we build a knowledge source consisting of multiple web documents. We expand on both steps below:

\paragraph{Entity Extraction.} 
As an initial filtering step, we use GPT-3.5 to identify proper nouns from the non-toxic and English user turns in WildChat. 
However, GPT-3.5 occasionally tags common nouns as proper nouns; we improve accuracy by additionally using GPT-4o to verify each identified proper noun. Finally, we prompt GPT-4o to identify and remove proper nouns with multiple meanings.

\paragraph{Building Knowledge Source.} Next, we build a knowledge source for each extracted entity. Prior works like the \factscore benchmark~\cite{min2023factscore} exclusively rely on Wikipedia for this step. However, as shown in our work, only a fraction of real world entities have corresponding Wikipedia pages. For \benchmark, we use commercial search engines to extract relevant web documents and build our knowledge source. 

We use the Google Custom Search JSON API\footnote{https://developers.google.com/custom-search/v1/overview} to collect the top 10 web page results using each entity as the search keyword. We scrape these web pages, exclude invalid and paywalled URLs and clean up the HTML and CSS tags in the remaining corpus. Appendix~\ref{sec:data-collection} outlines these steps in detail. After these cleaning steps, some entities are left with no web pages indicating insufficient information is available for these on the web. We remove such entities from our dataset.

\begin{figure}
    \centering
    \includegraphics[width=0.59\textwidth]{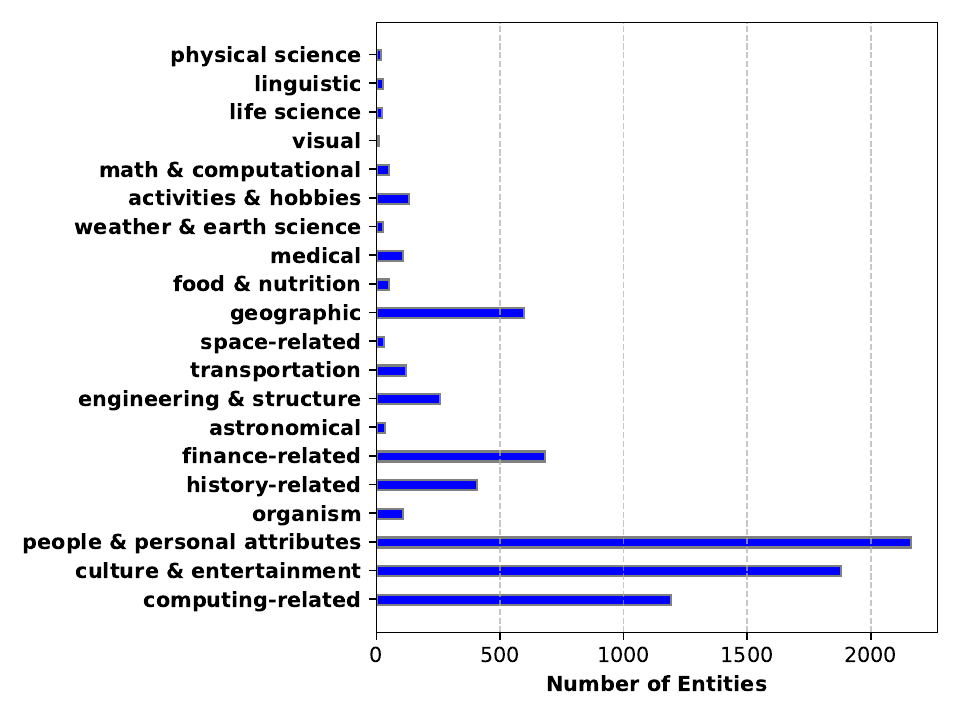}
    \includegraphics[width=0.39\textwidth]{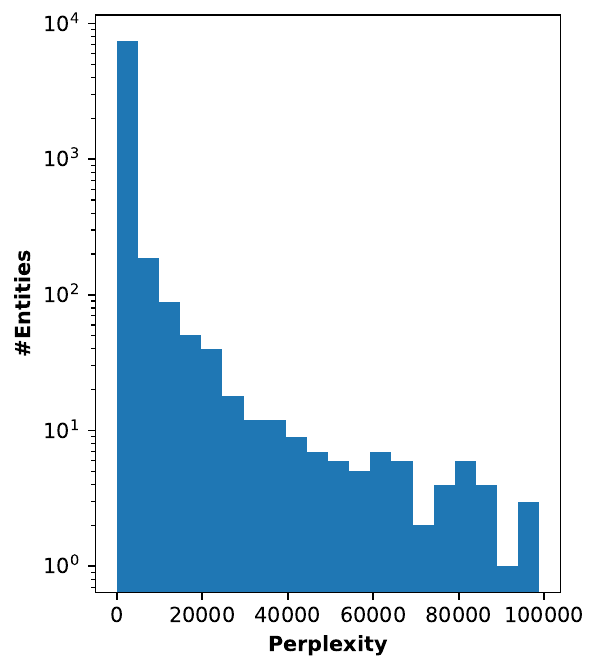}
    \caption{\label{fig:entity-analysis}\textbf{Left}: Distribution over domains of entities. \textbf{Right}: Distribution over perplexities of entities.}
    \vspace{-0.5cm}
\end{figure}

\subsection{\label{sec:benchmark-stats}Dataset Statistics}
Our data collection results in 7,917 entities. On average, each entity is associated with 6.08$_{\pm 2.83}$ web pages, collectively consisting of 69910$_{\pm 462671.91}$ tokens. \textbf{Notably, 52\% of the entities in \benchmark do not have Wikipedia pages as part of their knowledge source.} This highlights a key difference with existing benchmarks  that evaluate exclusively on Wikipedia entities that are likely to be more prominently featured in training data hence less challenging for LLMs. 

We manually inspect the evidence web pages for a random set of 100 entities. Our human evaluation confirms that all these entities are unambiguous and surface relevant web pages; we posit that searching for information on a single unambiguous entity is straightforward. However, not every web page is directly relevant to the entity, and sometimes the same information is repeated multiple times across different pages. 

\paragraph{Entity Domains.} A desiderata for our \benchmark benchmark is that it cover a diverse set of domains. Here, we analyze the domain distribution of the entities in our benchmark. 

We adopt the domain categories proposed by Wolfram Language for their natural language search engine \cite{WolframEntity} to classify the entities in \benchmark. We use the GPT-4o model to perform the category classification automatically. We ensure that each entity is associated with a single domain category; if multiple categories are relevant, we instruct GPT-4o to select the closest match. 

\Cref{fig:entity-analysis} (left) shows the distribution of entities across different domains. We find that the ``people \& personal attributes'', ``culture \& entertainment'', and ``computing-related'' domains have the most entities, whereas the science-related domains have the fewest entities. We also list example entities from each domain in \Cref{tab:entity-examples}. From our preliminary analysis, Wikipedia has the best coverage of science-related and history-related entities but has less coverage of newer or more niche entities.

\paragraph{Rare Entities.} Next, we analyze the frequency of the extracted entities. We approximate the an entity's frequency by computing its perplexity, without any surrounding context, using Llama-3-8B. The distribution of entity perplexity is shown in \Cref{fig:entity-analysis} (right). Note that higher perplexity indicates that the entity is rarer. While most entities fall in a perplexity range of 0-5000, the distribution exhibits a long tail, including extremely rare entities with a perplexity around $10^{4}$.

\subsection{Evaluation Pipeline}
\paragraph{Automatic Fact-Checking.} We apply the pipeline described in \factscore~\cite{min2023factscore} to perform automatic fact-checking. This pipeline consists of three steps: (1) decomposing a long-form generation into a set of atomic claims, (2) retrieving a number of passages for each atomic claim, and (3) verifying if each atomic claim is entailed by the retrieved passages.

\paragraph{Metrics.} We define two metrics. The first metric, \wildfactscore, computes the percentage of atomic facts supported by the knowledge source. Formally, let $M$ be a model to be evaluated, $x$ be a prompt asking for information about an entity of interest, and $S$ be a knowledge source. Given a response $y = M(x)$ and $A_y$, a list of atomic facts in $y$, \wildfactscore for an entity is calculated as \(\frac{1}{|A_y|} \sum_{a \in A_y} \mathbbm{1}[a \text{ is supported by } S]\), and this score is only computed for and averaged over the generations where $M$ does not abstain from providing information.

\wildfactscore has the following limitations: (1) many atomic facts may be trivially true in practice, leading to high \wildfactscore values, and (2) \wildfactscore does not account for abstention. Therefore, if a model abstains for most entities and gets the rest of the generations fully correct, \wildfactscore will be 100\%.
To address these limitations, we define a second metric, \wildfactscoreStrict, which assigns a score of 1 if all atomic facts about an entity are correct, and 0 if any atomic fact is wrong or the model abstains.

\begin{table}[!t]
\centering
\caption{\label{tab:entity-examples}Example entities from 20 domains classified by GPT-4o. The entities that do not have associated Wikipedia pages are bolded and colored in orange.}
\begin{tabular}{@{}ll@{}}
\toprule
Category & Examples \\ \midrule
geographic & Nigeria, Dominican Republic, Chugach State Park, Sichuan Province \\
astronomical & Hoag's Object, Drake Equation, Olbers' Paradox, {\color{orange} \textbf{GW190425}} \\
space-related & {\color{orange} \textbf{Satellite A105S2716}}, Giant Magellan Telescope, Outer Space Treaty \\
weather \& earth science & {\color{orange}\textbf{WeatherScraper}}, Hurricane Katrina, Carrington Event \\
transportation & Peugeot 508, {\color{orange}\textbf{Carnival Cruise Line}}, {\color{orange}\textbf{Jeyam Driving School}} \\
engineering \& structure & IIT Kharagpur, King Abdulaziz University, {\color{orange}\textbf{Aspen Plus}} \\
culture \& entertainment & {\color{orange}\textbf{CharlieIntel}}, {\color{orange}\textbf{Dentaverse podcast}}, Super Mario Odyssey \\
activities \& hobbies & Toledo Rockets, Fukuoka SoftBank Hawks, Boston Celtics \\
finance-related & {\color{orange}\textbf{Digital Markets Unit}}, {\color{orange}\textbf{Binance Smart Chain}}, {\color{orange} \textbf{Refinitiv Eikon}} \\
food \& nutrition & Mother Energy Drink, {\color{orange}\textbf{BiOptimizers}}, {\color{orange}\textbf{Lunatic Fridge IPA}} \\
people \& personal attributes & Dr. Gabor Mate, {\color{orange}\textbf{Harry Fayt}}, Tom Vilsack \\
history-related & Vietnam War, Achaemenid Empire, Salic Law \\
linguistic & Biblical Hebrew, Glagolitic alphabet, {\color{orange}\textbf{Princeton Language Institute}} \\
physical science & {\color{orange}\textbf{Bruker Vertex 70V}}, Oganesson, {\color{orange}\textbf{SciTechDaily}} \\
life science & {\color{orange}\textbf{Monocle3}}, {\color{orange}\textbf{Novaseq6000}}, SKBR3, HindIII \\
medical & {\color{orange}\textbf{Mucinex}}, Crohn's disease, {\color{orange}\textbf{Liquvida Naples}}, Lovenox \\
organism & Adalatherium, Bacillus cereus, Titanoboa, Caenorhabditis elegans \\
math \& computational & WeairePhelan, {\color{orange}\textbf{mysticalnumbers.com}}, NelderMead algorithm \\
computing-related & {\color{orange}\textbf{AHKCvJoyInterface}}, {\color{orange}\textbf{StructGPT}}, ProjectLibre, GraphQL \\
visual & {\color{orange}\textbf{Drawings.archicgi.com}}, {\color{orange}\textbf{Leica M10}}, {\color{orange}\textbf{Canon EOS R3}} \\ \bottomrule
\end{tabular}
\vspace{-0.35cm}
\end{table}

\begin{table}[t]
\centering
\caption{\label{tab:models}LLMs evaluated on \benchmark. \#Params indicates the number of parameters. Open denotes whether model weights are publicly available. Retrieval denotes whether a model leverages web search for its responses. With retrieval, a model's knowledge is considered up-to-date.}
\begin{tabular}{lrccc}
\toprule
Model & \#Params & Open & Retrieval & Knowledge Cutoff \\ \midrule
Llama-3-8B & 8B & \cmark & \xmark & 2023/03 \\
Llama-3-70B & 70B & \cmark & \xmark & 2023/12 \\
Mistral-7B & 7B & \cmark & \xmark & 2023/12 \\
Mixtral-8x7B & 56B & \cmark & \xmark & 2023/12 \\
Mixtral-8x22B & 176B & \cmark & \xmark & 2024/04 \\
Command R & 35B & \cmark & \cmark & up-to-date \\
Command R+ & 104B & \cmark & \cmark & up-to-date \\
Gemini 1.5 Flash & - & \xmark & \xmark & 2023/11 \\
Gemini 1.5 Pro & - & \xmark & \xmark & 2023/11 \\
Claude 3 Haiku & - & \xmark & \xmark & 2023/08 \\
Claude 3 Opus & - & \xmark & \xmark & 2023/08 \\
GPT-3.5 & - & \xmark & \xmark & 2021/09 \\
GPT-4o & - & \xmark & \xmark & 2023/10 \\
Sonar-Small & 8B & \xmark & \cmark & up-to-date \\
Sonar-Large & 70B & \xmark & \cmark & up-to-date \\
\bottomrule
\end{tabular}
\end{table}

\section{\label{sec:experimental_setup}Experimental Setup}
\paragraph{Models.} We evaluate 15 state-of-the-art LLMs: Llama-3-8B and Llama-3-70B \cite{llama3modelcard}; Mistral-7B, Mixtral-8x7B, and Mixtral-8x22B \cite{jiang2023mistral}; Command R and Command R+ (with web search enabled) \cite{commandr}; Gemini 1.5 Flash and Gemini 1.5 Pro \cite{team2023gemini}; Claude 3 Haiku (claude-3-haiku-20240307) and Claude 3 Opus (claude-3-opus-20240229) \cite{claude3}; GPT-3.5 (gpt-3.5-turbo-0125) and GPT-4o (gpt-4o-2024-05-13) \cite{achiam2023gpt}; and Perplexity AI models \footnote{\url{https://www.perplexity.ai/}} (llama-3-sonar-small-32k-online and llama-3-large-small-32k-online). These models vary in size, architecture (single vs. mixture-of-experts), degree of openness (open vs closed weights), and knowledge cutoff dates. More details are shown in \Cref{tab:models}.

\paragraph{Inference.} To evaluate the LLMs' knowledge about each entity, we prompt them with the question, ``In a paragraph, could you tell me what you know about [entity]?'' For a fair comparison, we limit the generation to 512 tokens, except for Command R and Command R+, which are given 1,024 tokens due to their shorter tokens. When generating responses, we set the sampling temperature to 1.

\paragraph{Model Abstention.} Different models use slightly different phrasing to indicate their refusal to generate facts about the queried entity. We manually reviewed 100 generations from each model to compile a list of such phrases. During evaluation, we identify responses containing any of these phrases as abstentions. These phrases are listed in \Cref{sec:model-abs}.

\begin{table}[]
\centering
\caption{\label{tab:generation_stats}Statistics of model responses on \benchmark. \%Responding indicates the percentage of generations that do not abstain. \#Facts/Res indicates the number of atomic facts per response. The number of tokens is computed with the word tokenizer in NLTK~\cite{bird-loper-2004-nltk}.}
\begin{tabular}{lccc}
\toprule
Model & Tokens & \%Responding & \#Facts/Res \\ \midrule
Llama-3-8B & 149.04 & 91.03 & 27.40 \\
Llama-3-70B & 160.53 & 88.07 & 30.27 \\
Mistral-7B & 170.49 & 99.10 & 30.08 \\
Mixtral-8x7B & 200.26 & 96.71 & 35.58 \\
Mixtral-8x22B & 137.71 & 95.25 & 24.85 \\
Command R & 143.16 & 99.57 & 25.77 \\
Command R+ & 147.83 & 97.22 & 27.15 \\
Gemini 1.5 Flash & 110.04 & 87.01 & 22.79 \\
Gemini 1.5 Pro & 114.18 & 88.07 & 30.27 \\
Claude 3 Haiku & 143.61 & 72.59 & 27.12 \\
Claude 3 Opus & 146.28 & 84.22 & 28.40 \\
GPT-3.5 & 114.20 & 96.87 & 21.80 \\
GPT-4o & 147.26 & 95.49 & 26.71 \\
Sonar-Small & 258.06 & 97.86 & 44.92 \\	
Sonar-Large & 233.43 & 97.66 & 40.87 \\ \bottomrule
\end{tabular}
\end{table}

\begin{figure}[h]
    \centering
    \includegraphics[width=0.9\textwidth]{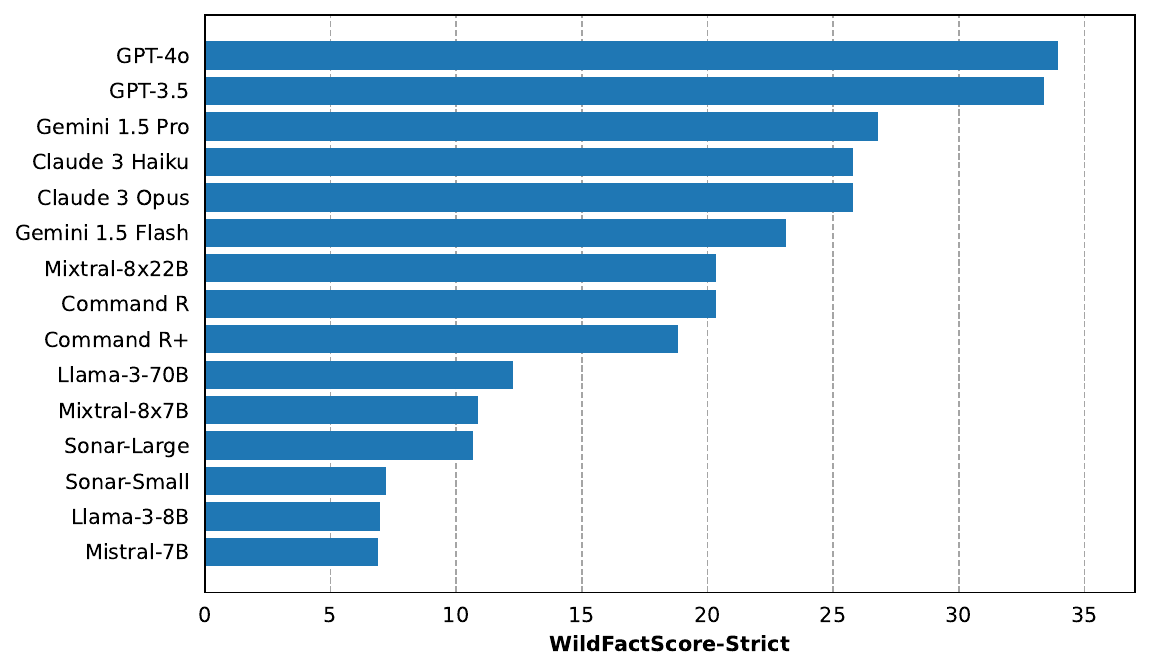}
    \caption{\label{fig:factaccuracy}\wildfactscoreStrict of 15 models. Models are sorted based on their \wildfactscoreStrict.}
\end{figure}

\begin{table}[h]
\centering
\caption{\label{tab:factscore}\wildfactscore of 15 models. We show \wildfactscore for all entities combined and broken down by domain. Models are sorted based on \wildfactscore for all entities combined. The top LLM in each domain is bolded.}
\begin{tabular}{lcccccc}
\toprule
Model & People & Culture & Geographic & Computing & Finance & All \\ \midrule
Claude 3 Haiku & \textbf{94.39} & 93.81 & 92.82 & 93.55 & \textbf{91.88} & \textbf{93.40} \\
GPT-4o & 88.16 & \textbf{94.55} & \textbf{94.05} & \textbf{93.86} & 91.10 & 92.33 \\
Claude 3 Opus & 90.89 & 93.40 & 92.29 & 92.29 & 89.86 & 92.10 \\
Gemini 1.5 Pro & 89.99 & 91.95 & 93.01 & 92.42 & 90.29 & 91.61 \\
GPT-3.5 & 80.11 & 92.58 & 92.18 & 91.82 & 89.38 & 88.80 \\
Llama-3-70B & 81.47 & 88.70 & 89.57 & 89.36 & 86.21 & 87.04 \\
Gemini 1.5 Flash & 78.21 & 86.59 & 91.49 & 91.28 & 88.36 & 86.65 \\
Mixtral-8x22B & 75.57 & 89.74 & 90.37 & 90.92 & 87.61 & 86.13 \\
Command R+ & 78.13 & 89.67 & 88.34 & 88.19 & 84.85 & 85.60 \\					
Command R & 76.84&	89.57	& 89.83	& 88.86	& 84.48	& 85.40\\
Sonar-Large & 78.36 & 88.38 & 87.57 & 87.52 & 85.01 & 85.19 \\
Sonar-Small & 77.47&86.33&	85.41&	85.80	&83.88&	83.53 \\
Mixtral-8x7B & 71.55 & 85.92 & 88.15 & 88.99 & 85.18 & 83.06 \\
Llama-3-8B & 67.10 & 78.95 & 85.96 & 85.12 & 81.27 & 78.60 \\
Mistral-7B & 58.87 & 76.27 & 84.70 & 86.02 & 81.57 & 75.07 \\ \bottomrule
\end{tabular}
\end{table}

\section{Results}
\paragraph{Response statistics across LLMs.}
We summarize response statistics for LLMs in \Cref{tab:generation_stats}. We find considerable variance in the length of the generated responses across different models; the Sonar models generate the longest responses and have the highest number of atomic facts, while GPT-3.5 and Gemini 1.5 Flash generate the shortest responses and have the lowest number of atomic facts. We also find that Claude 3 Haiku abstains the most, with only $72.59\%$ responding rate.
As expected, the retrieval-augmented Command R models abstains the least. In general, we observe that RAG models generate significantly longer responses and abstain less compared to the models without retrieval.

\paragraph{Comparing \wildfactscoreStrict of LLMs.} \Cref{fig:factaccuracy} presents the \wildfactscoreStrict scores for the 15 models we evaluate. We find that GPT-4o and GPT-3.5 perform comparably, achieving the highest \wildfactscoreStrict and outperforming the next best models by a significant margin of six absolute points. Among the open-weight models, Mixtral 8x22B achieves the best performance. Notably, even though Command R and Sonar models have access to web search, they still underperform the best models that lack retrieval. In fact, Sonar-Large performs worse than Llama-3-70B even though it uses Llama-3-70B as the base model and adds retrieval functionality. This mirrors findings from earlier work \cite{min2023factscore} that reported low performance of retrieval-augmented models. Even within a model family, larger models do not necessarily perform better.

We also find that despite claims of superior capabilities, GPT-4o and Claude 3 Opus exhibit factuality on par with GPT-3.5 and Claude 3 Haiku, respectively. We note that Gemini 1.5 Pro and the Claude models abstain substantially more often than GPT-4o and GPT-3.5, which further penalizes the former's \wildfactscoreStrict scores. Finally, closed-weight models consistently outperform open-weight models, leaving large room for improvement within the open-source research community.

\paragraph{Comparing \wildfactscore of LLMs} Next, we compare the \wildfactscore of different LLMs. We report \wildfactscore for all entity category types with more than 500 entities in addition to overall score. \Cref{tab:factscore} summarizes the results. First, we observe that \wildfactscore and \wildfactscoreStrict report substantially different trends. In particular, \textbf{we observe that Claude 3 Haiku outperforms all other models according to \wildfactscore, including the best performing GPT-4o models from Figure~\ref{fig:factaccuracy}.} One source of this discrepancy stems from model abstention---\wildfactscore is only calculated on generations where models do not abstain and therefore does not penalize models for high abstention rates. From \Cref{tab:generation_stats}, we already know that Claude 3 Haiku abstains for the highest percentage of queried entities. 

Analyzing different domains, we find that geographic and computing are relatively easier domains: 7 out of 15 models achieve over 90\% \wildfactscore in these domains. In contrast, people and finance are more challenging domains, with only two models and three models reporting over 90\% \wildfactscore in these domains, respectively. We also observe that although Claude 3 Haiku outperforms GPT-4o, this is primarily due to higher performance on the majority people category in \benchmark. By extracting entities in the wild from across various categories, our dataset provides a more fine-grained view into LLM performances, which is missing in similar prior benchmarks \cite{min2023factscore}.

\begin{figure}
    \centering
    \includegraphics[width=\linewidth]{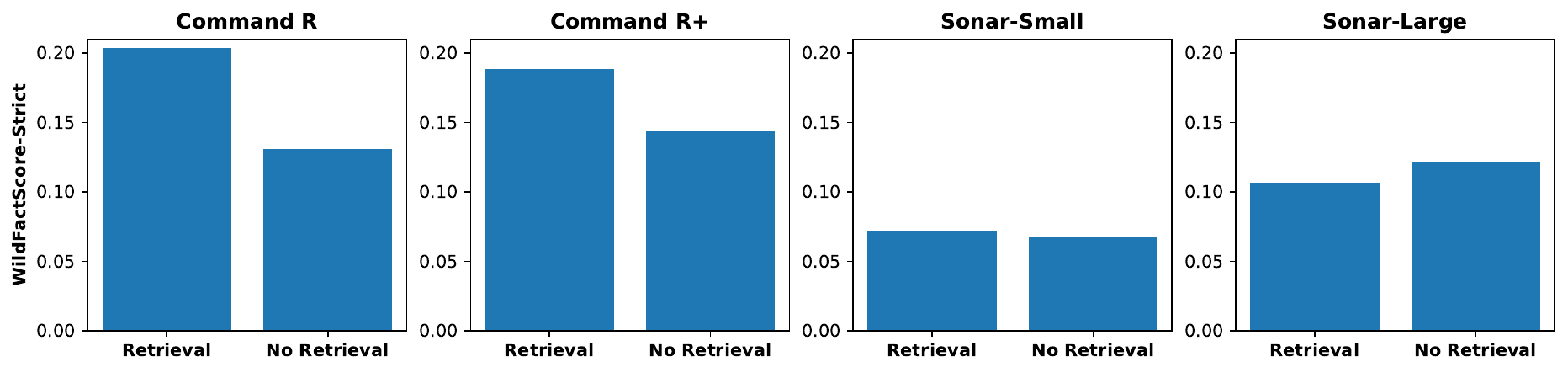}
    \caption{\label{fig:retrieval-analysis}Comparison of \wildfactscoreStrict with and without retrieval in RAG models.}
\end{figure}

\section{Analysis}
\paragraph{Do models hallucinate less with retrieval?}
We investigate the extent to which retrieval aids in reducing model hallucinations. To do this, we apply the same fact-checking pipeline to the RAG models with web retrieval disabled. Our findings, depicted in Figure~\ref{fig:retrieval-analysis}, include the \wildfactscoreStrict for four RAG models: Command R, Command R+, Sonar-Small, and Sonar-Large. We find that, without access to the web, all the models except Sonar-Large exhibit more frequent hallucinations. Interestingly, Sonar-Large hallucinates less in the absence of web retrieval.

\begin{figure}
    \centering
    \includegraphics[width=\textwidth]{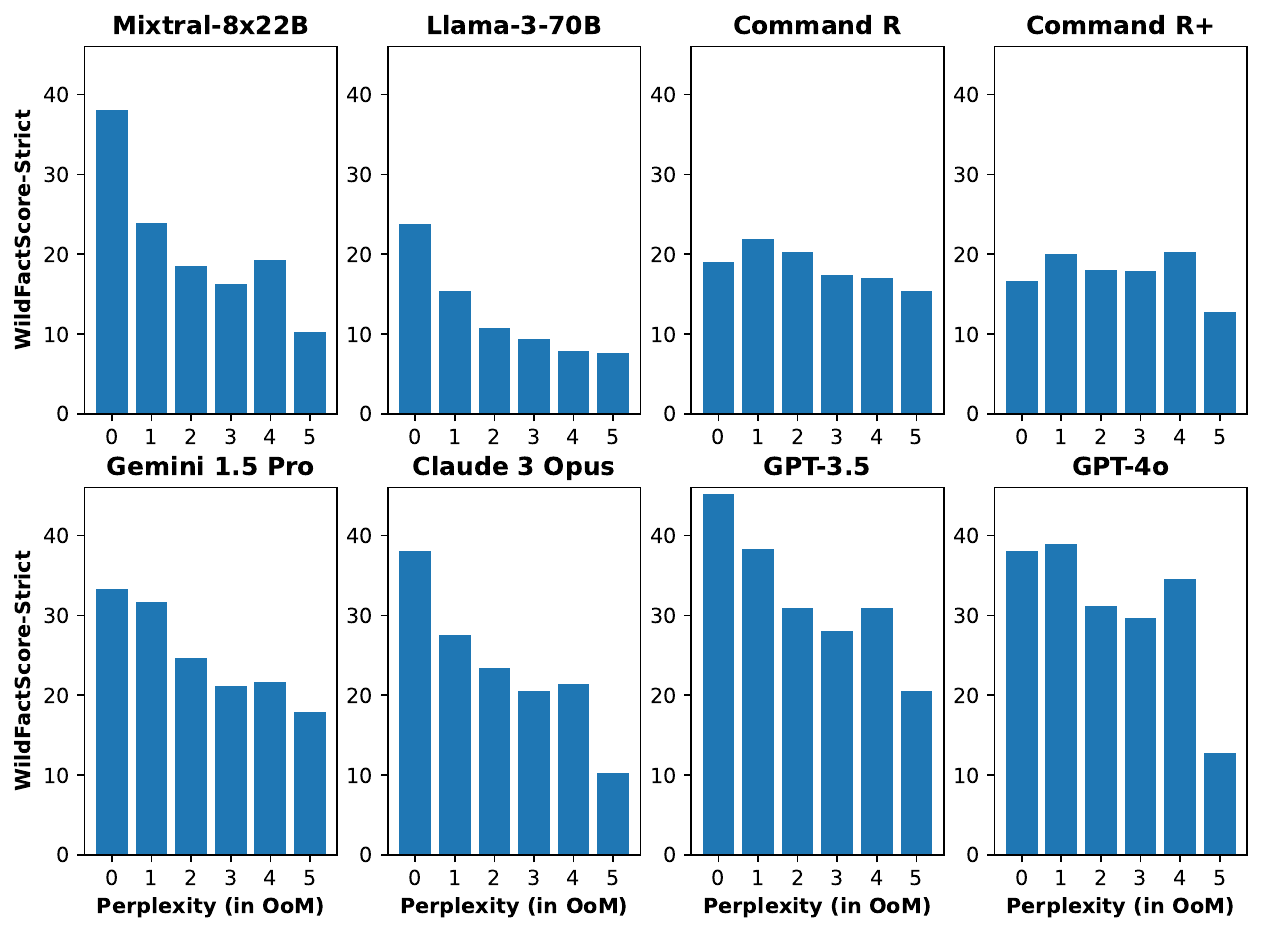}
    \caption{\label{fig:perplexity-analysis}\wildfactscoreStrict for eight LLMs on entities grouped by perplexity values.}
\end{figure}

\paragraph{Do models hallucinate more on rare entities?}
We evaluate factuality of LLMs as a function of entity frequency using the perplexity values in \Cref{sec:benchmark-stats}. We group the entities into 6 groups by the order of magnitude (OoM) of their perplexity: an OoM of 0 corresponds to the most frequent entities, while an OoM of 5 corresponds to the least frequent entities. We present the results of eight LLMs in \Cref{fig:perplexity-analysis}. Regardless of model size, models without retrieval report significantly decreased performance on rarer entities. Mixtral-8x22B and Claude 3 Opus show the largest drop in \wildfactscoreStrict. Conversely, models with retrieval are more robust to rare entities.

\begin{figure}
    \centering
    \includegraphics[width=\textwidth]{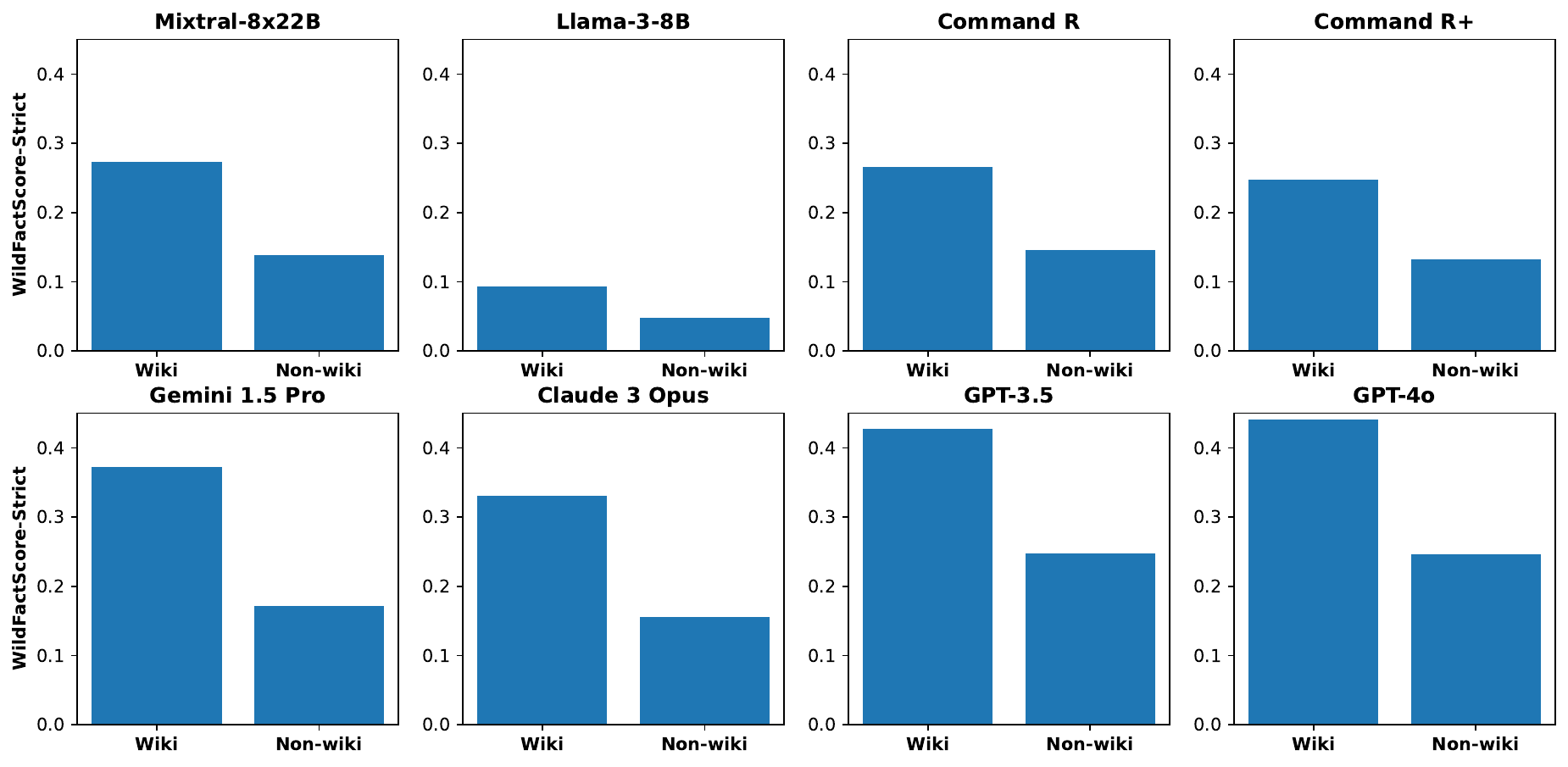}
    \caption{\label{fig:wiki-analysis}\wildfactscoreStrict broken down by whether the entities have Wikipedia pages.}
\end{figure}

\paragraph{Do models hallucinate more on non-Wikipedia knowledge?}
We also compare the factuality of LLMs on entities that have Wikipedia pages with those that do not. 
The results for eight LLMs are presented in Figure~\ref{fig:wiki-analysis}. We observe a significant decrease in \wildfactscoreStrict when recalling knowledge from sources other than Wikipedia for all eight models, with GPT-3.5 and GPT-4o exhibiting the largest drop. Interestingly, even though Command R and Command R+ models perform web searches, they also exhibit lower factual accuracy when generating information from non-Wiki sources. 

\paragraph{Do different LLMs hallucinate on similar entities?} \Cref{tab:generation_stats} shows that Claude and Gemini models generate facts about a substantially lower fraction of entities ($<90\%$) compared to the GPT models ($>95\%$). We hypothesize that some entities are challenging for all LLMs, and the former models are able to improve their overall \wildfactscore (\Cref{tab:factscore}) by being more conservative in responding.

\begin{table}[!t]
    \centering
    \caption{\label{tab:factscore-abstain}\wildfactscore and \wildfactscoreStrict on the subset of entities for which none of the top 5 models abstain. We observe that, contrary to the results in Table~\ref{tab:factscore}, GPT-4o outperforms all other models according to both metrics in this more apples-to-apples comparison on the exact same set of entities.}
    \small
    \begin{tabular}{lccccc}
    \toprule
        & Claude-3-Haiku & Claude-3-Opus & Gemini-Pro-1.5 & GPT-3.5 & GPT-4o \\ \midrule
         \wildfactscore & 93.7 & 93.4 & 92.9 & 93.7 & 95.1\\
         \wildfactscoreStrict & 33.2 & 29.4 & 34.5 & 39.0 & 39.7 \\
    \bottomrule
    \end{tabular}
\end{table}

To facilitate a more apples-to-apples comparison, we report \wildfactscore on a subset of entities for which all LLMs generate a response in \Cref{tab:factscore-abstain}. On these 5,891 entities, we observe that the top four models (GPT-3.5, GPT-4o, Gemini 1.5 Pro, Claude 3 Opus and Claude 3 Haiku) report similar performances. In contrast to \Cref{tab:factscore}, GPT-4o reports better \wildfactscore performance than Claude 3 Haiku. \textbf{This shows that the higher overall performance of Claude 3 Haiku can be attributed to its ability to abstain from generating facts about the more challenging subset of entities.}

\section{Related Work}
\paragraph{LLM Hallucinations} Hallucination has long been studied in traditional text generation settings like summarization and machine translation \cite{maynez-etal-2020-faithfulness, durmus-etal-2020-feqa, goyal2020evaluating}. In these scenarios, hallucinations are generations that contradict the source document. Nowadays, LLMs are increasingly used in closed-book settings where they rely on their parametric knowledge to answer queries and often hallucinate incorrect world knowledge or reasoning \cite{zhu2024halueval, zhao2024felm}. Recent works have characterized hallucinations in both settings \cite{zhao2024felm, li-etal-2023-halueval, malaviya2024expertqa}, covering diverse domains including scientific questions \cite{malaviya2024expertqa}, questions in the wild from users \cite{li-etal-2023-halueval}, and domain-specific errors in dialog datasets \cite{tang2024tofueval}. The most closely related dataset to our work is \factscore \cite{min2023factscore} which proposes a framework to evaluate LLM hallucinations in people-centric generations. Our work expands the scope of the benchmark by extracting a diverse set of entities from real user queries and using a more comprehensive knowledge source than Wikipedia.

\paragraph{Detecting LLM errors} Research in hallucination detection falls into two categories; the first line of work uses a question generation and answering pipeline where questions are asked of both the evidence document and generated content and the answers are matched \cite{durmus-etal-2020-feqa, wang2020asking, fabbri-etal-2022-qafacteval}. This allows for the use of strong off-the-shelf question-answering models without additional training. The other line of work uses entailment models to evaluate factuality, with entailment judgments being sought either at the output \cite{maynez-etal-2020-faithfulness}, sentence \cite{laban2022summac} or dependency arc level \cite{goyal2020evaluating}. Recently, hallucination evaluation models have  added a retrieval module to detect errors in closed-book settings \cite{min2023factscore, mishra2024fine, gao2023enabling}. Finally, instead of task-specific models that may not generalize to diverse LLM generations, recent work increasingly leverages strong LLMs like GPT-3.5 to verify entailment of generated facts \cite{min2023factscore,kamoi2024evaluating}.

\section{Conclusions}
We introduce \benchmark, a benchmark designed to evaluate the factuality of LLMs using entities extracted from real-world user-chatbot interactions. By prompting LLMs to generate information about these entities, \benchmark provides an evaluation of LLM factuality on use-case-relevant knowledge. The generated descriptions are fact-checked against a variety of web sources beyond traditional Wikipedia articles, offering a realistic evaluation of LLM factuality.


\bibliography{neurips_2024}
\bibliographystyle{plain}

\newpage
\appendix
\begin{table}[b!]
    \centering
    \caption{\label{tab:factscore-analysis} Error analysis of the evaluation pipeline.}
    \begin{tabular}{lccc}
    \toprule
         & Claude 3 Haiku & GPT-4o & Llama-3-70B \\ \midrule
        Retrieval &11.11 &11.11 & 15.56\\
        Entailment &14.81 &\phantom{0}8.33& 11.11 \\ 
        Total &25.92&19.44& 26.67 \\\bottomrule
    \end{tabular}
\end{table}

\begin{table}[b!]
\caption{\label{tab:rag-analysis} Error analysis of retrieval-augmented models (RAGs).}
\begin{tabular}{c p{1.7cm}p{1.5cm}p{3cm}p{4.5cm}}
\toprule
Error \% & Error Type & Entity & Atomic Fact & Retrieved Contexts \\ \midrule
43.70 & Generation error & libgdx & LibGDX was created in 2009. & In the middle of 2009, Mario Zechner, the creator of libGDX, \textit{wanted} to write Android games and \textit{started} developing a framework called AFX (Android Effects) for this. \\
27.73 & Evaluation error & Canon EOS R3 & The touch screen has 4.2 million dots. & Fully articulated, 8.2cm, 4,200,000-dot touch screen \\
17.65 & Contradictory information on Web & Dongmei & Dongmei Zhang is a Deputy Managing Director. & (Source 1:) Dr. Dongmei Zhang is a Distinguished Scientist and Assistant Managing Director at Microsoft Research Asia. (Source 2:) Dongmei Zhang is a Distinguished Scientist and Deputy Managing Director at Microsoft Research Asia. \\
4.20 & Retrieval error & Marta Batmasian & Bark \& Brunch' raises money for the Tri-County Animal Rescue. & More than 125 pooch-loving advocates filled the outdoor courtyard for an inaugural event that raised money for Tri-County Animal Rescue. \\
3.36 & Unverifiable opinions & COBIT 2019 & COBIT 2019 offers best-practice guidance on cost management. & N/A
\\
0.84 & Outdated information & Lamb Dynamic Lights & The mod is incompatible with Sodium v0.5.X. & Github Issue: The mod wouldn't work with Sodium v0.5.0 \\ \toprule
\end{tabular}
\end{table}

\section{Analysis}
\paragraph{Validating Evaluation Pipeline.}
We apply the evaluation pipeline introduced in \cite{min2023factscore}, which demonstrates a high agreement rate between human judgments and the proposed evaluation method. We further validate this evaluation pipeline on \benchmark. There are two possible sources of errors in the evaluation pipeline: (1) an atomic fact may indeed be supported, but the retrieved contexts may not contain relevant information, and (2) the entailment prediction may be incorrect. To check for these errors, we randomly sample 50 entities and manually review all atomic facts identified as unsupported from three models: Claude 3 Haiku, GPT-4o, and Llama-3-70B. We then calculate the proportion of errors that are actual mistakes versus those introduced by the evaluation process. We summarize the results in Table~\ref{tab:factscore-analysis}. We find that 19\%-26\% of the atomic facts identified as unsupported are not actual mistakes. However, since the evaluation pipeline introduces similar proportions of mistakes across models, the models are roughly equally affected by these errors.

\begin{figure}[b!]
    \centering
    \footnotesize
    \framebox{
    \parbox{0.9\textwidth}{
    \textbf{Entity}:\newline
    University of Cambridge \smallskip \newline
    \textbf{Generation}:\newline
    The University of Cambridge is a public collegiate research university located in Cambridge, England. Founded in 1209, it is the world's third-oldest university in continuous operation. The university's reputation for outstanding academic achievement is known worldwide and its alumni include Nobel Prize winners,\u77e5\u540d thinkers, and transformative figures in several fields. Cambridge's many famous colleges and university buildings attract visitors from around the world. The university is divided into six schools: Humanities and Social Sciences, Clinical Medicine, Technology, Biological Sciences, the Humanities and Social Sciences, and Physical Sciences. Each school comprises several faculties, departments, and other institutions. Teaching and learning at Cambridge are highly personalised and conducted in small groups, and the university's extensive resources include libraries and museums rich in treasures. \noindent \smallskip \newline
    \textbf{Atomic Facts}:\newline
{\color{red} The University of Cambridge is a university.} \newline
The University of Cambridge is a public university. \newline
The University of Cambridge is a collegiate research university.\newline
{\color{red} The University of Cambridge is located in Cambridge, England.}\newline
It was founded in 1209.\newline
{\color{red} It is a university.}\newline
It is the world's third-oldest university.\newline
{\color{red} It is in continuous operation.}\newline
The university has a reputation for outstanding academic achievement.\newline
The university's reputation is known worldwide.\newline
The university's alumni include Nobel Prize winners.\newline
The university's alumni include known thinkers.\newline
The university's alumni include transformative figures.\newline
The university's alumni are in several fields.\newline
Cambridge has many famous colleges.\newline
Cambridge has many famous university buildings.\newline
{\color{blue} Cambridge's colleges attract visitors from around the world.}\newline
{\color{blue}Cambridge's university buildings attract visitors from around the world.}\newline
{\color{blue} Cambridge's colleges and university buildings attract visitors from around the world.}\newline
The university is divided into six schools.\newline
The six schools are: Humanities and Social Sciences, Clinical Medicine, Technology, Biological Sciences, the Humanities and Social Sciences, and Physical Sciences.\newline
{\color{red} Each school comprises faculties.}\newline
{\color{red} Each school comprises departments.}\newline
{\color{red} Each school comprises other institutions.}\newline
{\color{blue}Teaching at Cambridge is highly personalised.}\newline
{\color{blue}Learning at Cambridge is highly personalised.}\newline
{\color{blue}Teaching at Cambridge is conducted in small groups.}\newline
{\color{blue}Learning at Cambridge is conducted in small groups.}\newline
Cambridge University has extensive resources.\newline
Cambridge University's resources include libraries.\newline
Cambridge University's resources include museums.\newline
Cambridge University's libraries are rich in treasures.\newline
Cambridge University's museums are rich in treasures.\newline
    }
    }
    \caption{A list of atomic facts decomposed from an LLM generation. Facts colored in red are trivially true, and facts colored in blue are repetitive facts.}
    \label{fig:atomic-facts}
\end{figure}

\paragraph{Analysis of RAGs.}
RAG models are designed with the goal of reducing hallucinations. Counterintuitively, they appear to have a similar amount of hallucinations as models without retrieval. To better understand the behaviors of RAGs, we analyze the hallucinations from these models. Specifically, we study whether the hallucinations are a result of incorrect retrieval or weak language model capabilities. We inspect the same 50 entities used in the previous analysis and examine a total of 119 unsupported atomic facts decomposed from generations by the Command R and Command R+ models.\footnote{We analyze Cohere models but not Perplexity AI models because the latter do not provide the retrieved contexts.} The results are summarized in Table~\ref{tab:rag-analysis}. Our findings show that 43.7\% of errors come from model generation, where correct contexts were retrieved, but the generation did not faithfully follow those contexts. Only 4.2\% of errors were due to incorrect retrieval, where irrelevant information was retrieved. The evaluation process also introduced 27\% of errors, a proportion similar to that observed in other models. Another significant source of error was contradictory information found on the web.

\paragraph{Qualitative Analysis of Trivial Atomic Facts.} \wildfactscore often suffers from trivial atomic facts, inflating its results. We illustrate examples of such trivial atomic facts in Figure~\ref{fig:atomic-facts}. A significant number of atomic facts are trivially true; for example, stating that "the University of Cambridge is a university." Additionally, many atomic facts are repetitive. For instance, the atomic facts highlighted in blue express the same meaning.

\section{Detecting Model Abstention}
We use the code below to detect if a model abstains from responding.
\label{sec:model-abs}
\begin{lstlisting}[language=Python]
invalid_ppl_mentions = [
    "I could not find any information",
    "The search results do not provide",
    "There is no information",
    "There are no search results",
    "there are no provided search results",
    "not provided in the search results",
    "is not mentioned in the provided search results",
    "There seems to be a mistake in the question",
    "Not sources found",
    "No sources found",
    "Try a more general question",
    "Unfortunately,",
    "There doesn't seem to be",
    "There does not seem to be",
    "Please",
    "I do not",
    "I don't",
    "**No relevant",
    "I'm afraid",
    "I am afraid",
    "I apologize,",
    "I'm sorry",
    "I am sorry"
    "Sorry",
    "I am not familiar with",
    "I'm not familiar with",
]

def wildentities_ai_abstain_detect(generation):
    if "is not explicitly mentioned" in generation:
        return True
    if "is not mentioned" in generation:
        return True
    if "isn't mentioned" in generation:
        return True
    if "is not a widely" in generation:
        return True
    if "isn't a widely" in generation:
        return True
    if is_invalid_ppl(generation):
        return True
    if "provide more" in generation:
        return True
    valid_paras = []
    for para in generation.split("\n\n"):
        if is_invalid_paragraph_ppl(para):
            break
        valid_paras.append(para.strip())

    if len(valid_paras) == 0:
        return True
    else:
        return False
\end{lstlisting}

\section{Data Collection}
\label{sec:data-collection}
We include more details about the web scraping process. We use ScrapingBee to scrape these web pages with an ad blocker enabled. To mitigate noise in the web content, we perform the following cleaning steps: (1) we remove any web page with a status code not equal to 200, (2) we exclude web pages from platforms that require login to view content\footnote{twitter.com, youtube.com, instagram.com, reddit.com, facebook.com, amazon.com, linkedin.com}, (3) we exclude websites containing phrases like ``You're signed out'' or ``Log in to,'' and (4) we use BeautifulSoup\footnote{\url{https://www.crummy.com/software/BeautifulSoup/}} to remove HTML and CSS tags. After these cleaning steps, some entities may be left with no web pages. This indicates that there is not enough information available on the web for these entities, and we thus remove those entities from our dataset.

\section{Limitations}
\label{sec:limitations}
\paragraph{Language Scope}
Despite the diversity of entity types used in \benchmark, the current version focuses on English. This limitation excludes the evaluation of LLM factuality in other languages, which is critical for assessing the performance of LLMs in multilingual contexts, espceially considering that powerful LLMs have multilingual capabilities and are used in different languages~\cite{ouyang2023the}.

\paragraph{Bias in Source Data}
The data used in \benchmark is sourced from the WildChat dataset, which reflects the demographic distribution of its users. Consequently, any biases present in this user base are inherited by \benchmark. This includes potential biases in user demographics, interests, and interaction styles. Addressing this bias would require diversifying the data sources that represent a wider demographic spectrum.

\paragraph{Noise in Web Content} One significant challenge in our approach is the inherent noise in web content. Articles retrieved from the Internet may contain misinformation, outdated information, or biased perspectives~\cite{chalkiadakis2021rise,bhattacharjee2020disinformation}. This noise can adversely affect the accuracy of the fact-checking process, as incorrect or misleading information may be used to verify the LLM's generated descriptions.

\paragraph{Retrieval Errors} Retrieval errors represent another limitation in our approach. While we strive to retrieve articles relevant to each entity, the retrieval step might not always identify the most relevant and up-to-date knowledge sources. This limitation can lead to incomplete or inaccurate evaluations of the factuality of the LLMs' responses. Improving retrieval algorithms is essential for mitigating this issue.

\paragraph{Entailment Errors} The fact-checking process involves decomposing generated descriptions into atomic claims and verifying these claims against the retrieved web articles. This step can introduce entailment errors, which occur in two forms: false positives and false negatives. False positives arise when a correct atomic fact is incorrectly identified as wrong, while false negatives occur when an incorrect atomic fact is mistakenly identified as correct. These errors can harm the evaluation results.

\end{document}